\documentclass[11pt,a4paper]{article}
\usepackage[hyperref]{emnlp-ijcnlp-2019}
\usepackage{times}
\usepackage{latexsym}
\usepackage{graphicx}
\usepackage{amsmath,amssymb,amsfonts}

\usepackage{url}
\aclfinalcopy 

\title{Utilizing BERT Intermediate Layers for Aspect Based Sentiment Analysis and Natural Language Inference}

\author{Youwei Song, Jiahai Wang \thanks{~~The corresponding author.}, Zhiwei Liang, Zhiyue Liu, Tao Jiang \\
  School of Data and Computer Science \\
  Sun Yat-sen University \\
  Guangzhou, China \\
  {\tt \{songyw5,liangzhw25,liuzhy93,jiangt59\}@mail2.sysu.edu.cn} \\
  {\tt wangjiah@mail.sysu.edu.cn} \\
}

\date{}

\begin{document}
\maketitle
\begin{abstract}
Aspect based sentiment analysis aims to identify the sentimental tendency towards a given aspect in text.
Fine-tuning of pretrained BERT performs excellent on this task and achieves state-of-the-art performances.
Existing BERT-based works only utilize the last output layer of BERT and ignore the semantic knowledge in the intermediate layers.
This paper explores the potential of utilizing BERT intermediate layers to enhance the performance of fine-tuning of BERT.
To the best of our knowledge, no existing work has been done on this research.
To show the generality, we also apply this approach to a natural language inference task.
Experimental results demonstrate the effectiveness and generality of the proposed approach.
\end{abstract}

\section{Introduction}

Aspect based sentiment analysis (ABSA) is an important task in natural language
processing. It aims at collecting and analyzing
the opinions toward the targeted aspect in an entire text. In the past decade, ABSA has received great attention due to a wide range of applications
\cite{pang2008opinion,liu2012sentiment}. Aspect-level
(also mentioned as “target-level”) sentiment classification as a subtask of ABSA \cite{pang2008opinion} aims at judging the sentiment polarity for a given aspect.
For example, given a sentence \textit{``I hated their service, but their food was great''}, the sentiment polarities for the target \textit{``service''} and \textit{``food''} are negative and positive respectively.

Most of existing methods focus on designing
sophisticated deep learning models to mining the relation
between context and the targeted aspect.
Majumder et al., \shortcite{majumder2018iarm} adopt a memory network
architecture to incorporate the related information
of neighboring aspects. 
Fan et al., \shortcite{fan2018multi} combine the fine-grained and coarse-grained attention 
to make LSTM treasure the aspect-level interactions.
However, the biggest challenge in ABSA task is the shortage of training data, 
and these complex models did not lead to significant improvements in outcomes.

Pre-trained language models can leverage large amounts of unlabeled data to learn
the universal language representations, which provide an effective solution for the above problem.
Some of the most prominent examples are ELMo \cite{peters2018deep}, GPT \cite{radford2018improving} and BERT \cite{devlin2018bert}.
BERT is based on a multi-layer bidirectional Transformer, and is trained on plain text for
masked word prediction and next sentence prediction tasks.
The pre-trained BERT model can then be fine-tuned on downstream task with task-specific training data.
Sun et al., \shortcite{sun2019utilizing} utilize BERT for ABSA task by constructing a auxiliary sentences,
Xu et al., \shortcite{xu2019bert} propose a post-training approach for ABSA task, and
Liu et al., \shortcite{liu2019multi} combine multi-task learning and pre-trained BERT to improve the performance of various NLP tasks. 
However, these BERT-based studies follow the canonical way of fine-tuning: append just an additional output layer after BERT structure.
This fine-tuning approach ignores the rich semantic knowledge contained in the intermediate layers.
Due to the multi-layer structure of BERT, different layers capture different levels of representations for the specific task after fine-tuning.

This paper explores the potential of utilizing BERT intermediate layers for facilitating BERT fine-tuning.
On the basis of pre-trained BERT, we add an additional pooling module, design some pooling strategies for integrating the multi-layer representations of the classification token.
Then, we fine tune the pre-trained BERT model with this additional pooling module and achieve new state-of-the-art
results on ABSA task. 
Additional experiments on a large Natural Language Inference (NLI) task 
illustrate that our method can be easily applied to more NLP tasks with only a minor adjustment.

Main contributions of this paper can be summarized as follows:

\begin{enumerate}
\item It is the first to explore the potential of utilizing intermediate layers of BERT 
and we design two effective information pooling strategies to solve aspect based sentiment analysis task.
\item  Experimental results on ABSA datasets show that our
method is better than the vanilla BERT model and 
can boost other BERT-based models with a minor adjustment.
\item  Additional experiments on a large NLI dataset illustrate that our method 
has a certain degree of versatility, and can be easily applied to some other NLP tasks.
\end{enumerate}

\section{Methodology}


\subsection{Task description}

\paragraph{ABSA}  Given a sentence-apsect pair, ABSA aims at predicting the sentiment polarity (\emph{positive}, \emph{negative} or \emph{neural}) of the sentence over the aspect. 

\paragraph{NLI} Given a pair of sentences, the goal is to predict whether a sentence is an \emph{entailment}, \emph{contradiction}, or \emph{neutral} with respect to the other sentence.

\subsection{Utilizing Intermediate Layers: Pooling Module}

\begin{figure}[t]
\setlength{\belowcaptionskip}{-15pt}
\includegraphics[width=\linewidth]{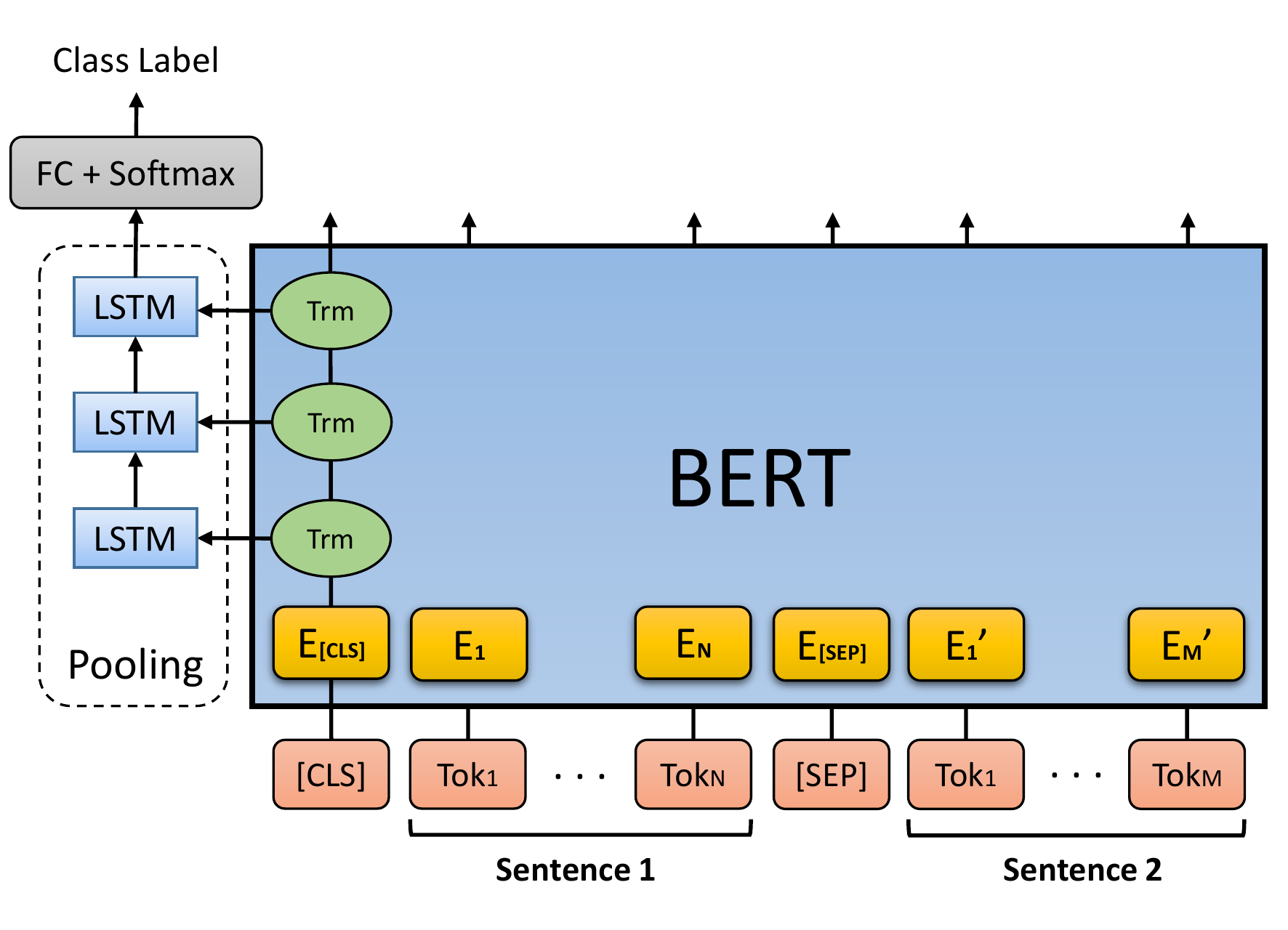}
  \caption{Overview of the proposed BERT-LSTM model. Pooling Module is responsible for connecting the intermediate representations obtained by Transformers of BERT.}
  \label{fig:model}
\end{figure}

Given the hidden states of the first token (i.e., {\tt [CLS]} token) 
$\mathbf{h}_{\tiny \textsc{CLS}} = \{h_{\tiny \textsc{CLS}}^1, h_{\tiny \textsc{CLS}}^2, ..., h_{\tiny \textsc{CLS}}^L\}$ from all $L$ intermediate layers. 
The canonical way of fine-tuning simply take the final one (i.e., $h_{\tiny \textsc{CLS}}^L$) for classification,
which may inevitably lead to information losing during fine-tuning.
We design two pooling strategies for utilizing $\mathbf{h}_{\tiny \textsc{CLS}}$: LSTM-Pooling and Attention-Pooling.
Accordingly, the models are named \textbf{BERT-LSTM} and \textbf{BERT-Attention}.
The overview of BERT-LSTM is shown in Figure \ref{fig:model}.
Similarly, BERT-Attention replaces the LSTM module with an attention module.

\paragraph{LSTM-Pooling}
Representation of the hidden states $\mathbf{h}_{\tiny \textsc{CLS}}$ is a special sequence: an abstract-to-specific sequence.
Since LSTM network is inherently suitable for processing sequential information,
we use a LSTM network to connect all intermediate representations of the {\tt [CLS]} token,
and the output of the last LSTM cell is used as the final representation.
Formally,
\begin{displaymath}
o = h_{\tiny \textsc{LSTM}}^L = \overrightarrow{\textsc{LSTM}}(h_{\tiny \textsc{CLS}}^i), i \in [1, L]
\end{displaymath}

\paragraph{Attention-Pooling} 
Intuitively, attention operation can learn the contribution of each $h_{\tiny \textsc{CLS}}^i$.
We use a dot-product attention module to dynamically combine all intermediates:
\begin{displaymath}
o = W_h^T softmax(\mathbf{q}\mathbf{h}_{\tiny \textsc{CLS}}^{T})\mathbf{h}_{\tiny \textsc{CLS}}
\end{displaymath}
where $W_h^T$ and $\mathbf{q}$ are learnable weights.

Finally, we pass the pooled output $o$ to a fully-connected layer for label prediction:
\begin{displaymath}
y = softmax(W_o^T o + b_o)
\end{displaymath}

\section{Experiments}

In this section, we present our methods for BERT-based model fine-tuning on three ABSA datasets.
To show the generality, we also conduct experiments on a large and popular NLI task.
We also apply the same strategy to existing state-of-the-art BERT-based models and demonstrate the effectiveness of our approaches.

\subsection{Datasets}

This section briefly describes three ABSA datasets and SNLI dataset.
Statistics of these datasets are shown in Table~\ref{tab:datasets}.

\paragraph{ABSA}
We use three popular datasets in ABSA task: Restaurant reviews and Laptop reviews from SemEval 2014 Task 4 \footnote{The detailed introduction of this task can be found at \url{http://alt.qcri.org/semeval2014/task4}.} \cite{pontiki2014semeval}, and ACL 14 Twitter dataset \cite{dong2014adaptive}.

\paragraph{SNLI}
The Stanford Natural Language Inference \cite{bowman2015large} dataset contains 570k human annotated hypothesis/premise pairs. 
This is the most widely used entailment dataset for natural language inference.

\begin{table}
    \setlength{\belowcaptionskip}{-15pt}
  \begin{center}
    \scalebox{0.9}{
    \begin{tabular}{l||c|c|c|c}
      \hline \bf Dataset & \#Train & \#Dev & \#Test   & \#Label \\ \hline \hline
      \multicolumn{5}{c}{Aspect Based Sentiment Analysis (ABSA)} \\ \hline
      Laptop & 2.1k & 0.2k & 0.6k & 3 \\ \hline \hline
      Restaurant & 3.2k & 0.4k & 1.1k & 3 \\ \hline
      Twitter & 5.6k & 0.6k & 0.7k & 3 \\ \hline
      \multicolumn{5}{c}{Natural Language Inference (NLI)} \\ \hline
      SNLI & 549k & 9.8k & 9.8k & 3 \\ \hline
    \end{tabular}
    }
  \end{center}
  \caption{Summary of the datasets. For ABSA dataset, we randomly chose 10\% of \#Train as \#Dev as there is no \#Dev in official dataset.}
  \label{tab:datasets}
\end{table}

\subsection{Experiment Settings}

All experiments are conducted with BERT$_{\tiny \textsc{BASE}}$ (uncased)
\footnote{\url{https://github.com/huggingface/pytorch-pretrained-BERT}.}
with different weights.
During training, the coefficient $\lambda$ of $\mathcal{L}_2$ regularization item is $10^{-5}$ and dropout rate is 0.1.
Adam optimizer \cite{kingma2014adam} with learning rate of 2e-5 is applied to update all the parameters.
The maximum number of epochs was set to 10 and 5 for ABSA and SNLI respectively.
In this paper, we use 10-fold cross-validation, which performs quite stable in ABSA datasets.

Since the sizes of ABSA datasets are small and there is no validation set, the results between two consecutive epochs may be significantly different.
In order to conduct fair and rigorous experiments, we use 10-fold cross-validation for ABSA task, which achieves quite stable results.
The final result is obtained as the average of 10 individual experiments.

The SNLI dataset is quite large, so we simply take the best-performing model on the development set for testing.

\subsection{Experiment-I: ABSA}

Since BERT outperforms previous non-BERT-based studies on ABSA task by a large margin, we are not going to compare our models with non-BERT-based models.
The 10-fold cross-validation results on ABSA datasets are presented in Table \ref{tbl:result_absa}.

\begin{table}
    \setlength{\belowcaptionskip}{-15pt}
    \centering
    \scalebox{0.65}{
        \begin{tabular}{l||c c|c c|c c}
        \hline
        {\bf Domain} & \multicolumn{2}{c}{\bf Laptop} & \multicolumn{2}{c}{\bf Restaurant} & \multicolumn{2}{c}{\bf Twitter} \\
        \hline
        {\bf Methods} & \bf{Acc.} & \bf{F1} & \bf{Acc.} & \bf{F1} & \bf{Acc.} & \bf{F1} \\
        \hline
        BERT$_{\tiny \textsc{BASE}}$ & 74.66 & 68.64 & 81.92 & 71.97 & 72.46 & 71.04 \\
        \textbf{BERT-LSTM} & \textbf{75.31} & \textbf{69.37} & 82.21 & 72.52 & 73.06 & 71.61 \\
        \textbf{BERT-Attention} & 75.16 & 68.76 & \textbf{82.38} & \textbf{73.22} & \textbf{73.35} & \textbf{71.88} \\
        \hline
        BERT-PT \cite{xu2019bert} & 76.27 & 70.66 & 84.53 & 75.33 & - & - \\
        \textbf{BERT-PT-LSTM} & 77.08 & 71.65 & \textbf{85.29} & \textbf{76.88} & - & - \\
        \textbf{BERT-PT-Attention} & \textbf{77.68} & \textbf{72.57} & 84.92 & 75.89 & - & - \\
        \hline
        \end{tabular}
    }
  \caption{Accuracy and macro-F1 (\%) for aspect based sentiment analysis on three popular datasets.}
\label{tbl:result_absa}
\end{table}

\begin{figure*}[ht]
\setlength{\belowcaptionskip}{-7pt}
\includegraphics[width=\linewidth]{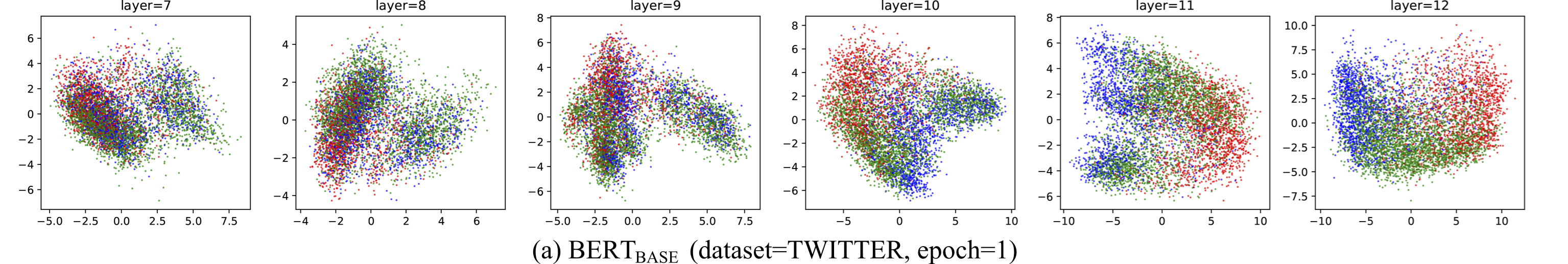}\vspace{2.5pt}
\includegraphics[width=\linewidth]{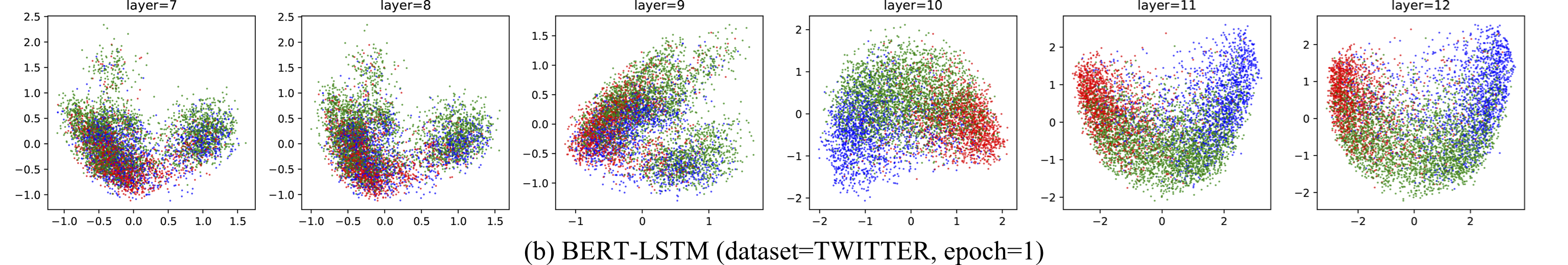}\vspace{2.5pt}
\includegraphics[width=\linewidth]{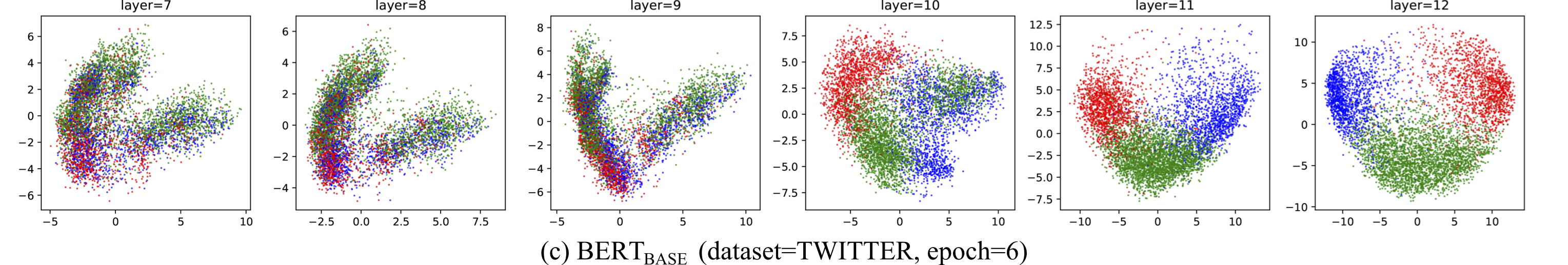}\vspace{2.5pt}
\includegraphics[width=\linewidth]{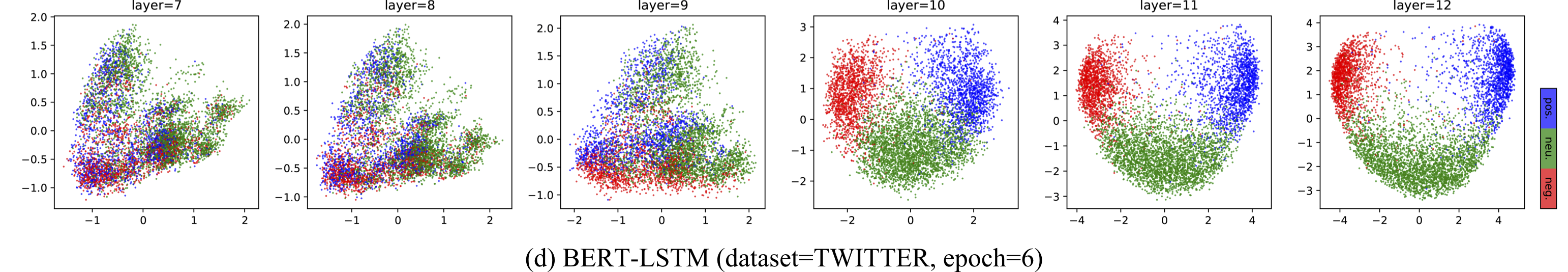}
  \caption{Visualization of BERT and BERT-LSTM on Twitter dataset with the last six intermediates layers of BERT at the end of the 1st and 6th epoch. Among the PCA results, (a) and (b) illustrate that BERT-LSTM converges faster than BERT after just one epoch, while (c) and (d) demonstrate that BERT-LSTM cluster each class of data more dense and discriminative than BERT after the model nearly converges.}
  \label{fig:absa_vis}
\end{figure*}

The BERT$_{\tiny \textsc{BASE}}$, BERT-LSTM and BERT-Attention are both initialized with pre-trained BERT$_{\tiny \textsc{BASE}}$ (uncased).
We observe that BERT-LSTM and BERT-Attention outperform vanilla BERT$_{\tiny \textsc{BASE}}$ model on all three datasets.
Moreover, BERT-LSTM and BERT-Attention have respective advantages on different datasets.
We suspect the reason is that Attention-Pooling and LSTM-Pooling perform differently during fine-tuning on different datasets.
Overall, our pooling strategies strongly boost the performance of BERT on these datasets.

The BERT-PT, BERT-PT-LSTM and BERT-PT-Attention are all initialized with post-trained BERT \cite{xu2019bert} weights
\footnote{Since our evaluation method is different from Xu et al., \shortcite{xu2019bert}, we post the results based on our experiment settings.}.
We can see that both BERT-PT-LSTM and BERT-PT-Attention outperform BERT-PT with a large margin on Laptop and Restaurant dataset
\footnote{Experiments are not conducted on Twitter dataset for Xu et el., \shortcite{xu2019bert} does not provide post-trained BERT weights on this dataset.}.
From the results, the conclusion that utilizing intermediate layers of BERT brings better results is still true.

\subsubsection{Visualization of Intermediate Layers}

In order to visualize how BERT-LSTM
\footnote{BERT-LSTM is more suitable for visualizing intermediate layers.}
benefits from sequential representations of intermediate layers, we use principal component analysis (PCA) to visualize the intermediate representations of {\tt [CLS]} token, shown in figure \ref{fig:absa_vis}.
There are three classes of the sentiment data, illustrated in blue, green and red, representing positive, neural and negative, respectively.
Since the task-specific information is mainly extracted from the last six layers of BERT,
we simply illustrate the last six layers.
It is easy to draw the conclusion that BERT-LSTM partitions different classes of data faster and more dense than vanilla BERT
under the same training epoch.



\subsection{Experiment-II: SNLI}

\begin{table}[t]
    \setlength{\belowcaptionskip}{-15pt}
  \begin{center}
    \small
    \begin{tabular}{l | c | c }\hline
     \bf Model &Dev& Test  \\ \hline
    GPT \cite{radford2018improving} &- & 89.9$^*$ \\
    \hline
    \citet{kim2018semantic} &- & 90.1$^*$ \\
        \hline
    BERT$_{\tiny \textsc{BASE}}$ &90.94 &90.66 \\
        \textbf{BERT-Attention} &91.12 &90.70 \\
        \textbf{BERT-LSTM} & \textbf{91.18} & \textbf{90.79} \\
    \hline
    MT-DNN \cite{liu2019multi} & 91.35 & \textbf{91.00} \\
        \textbf{MT-DNN-Attention} & 91.41 & 90.95 \\
        \textbf{MT-DNN-LSTM} &\textbf{91.50} & 90.91 \\
    \hline
    \end{tabular}
  \end{center}
  \caption{Classification accuracy (\%) for natural language
inference on SNLI dataset. Results with ``*'' are obtained from the official SNLI leaderboard (https://nlp.stanford.edu/projects/snli/).
  }
  \label{tab:snli}
\end{table}

To validate the generality of our method, we conduct experiment on SNLI dataset and apply same pooling strategies to currently state-of-the-art method \textbf{MT-DNN} \cite{liu2019multi}, which is also a BERT based model, named \textbf{MT-DNN-Attention} and \textbf{MT-DNN-LSTM}.


As shown in Table \ref{tab:snli}, the results were consistent with those on ABSA.
From the results, BERT-Attention and BERT-LSTM perform better than vanilla BERT$_{\tiny \textsc{BASE}}$.
Furthermore, MT-DNN-Attention and MT-DNN-LSTM outperform vanilla MT-DNN on Dev set, and are slightly inferior to vanilla MT-DNN on Test set.
As a whole, our pooling strategies generally improve the vanilla BERT-based model, which draws the same conclusion as on ABSA.

The gains seem to be small, but the improvements of the method are straightforwardly reasonable and the flexibility of our strategies makes it easier to apply to a variety of other tasks.

\section{Conclusion}

In this work, we explore the potential of utilizing BERT intermediate layers and propose two effective pooling strategies to enhance the performance of fine-tuning of BERT. Experimental results demonstrate the effectiveness and generality of the proposed approach.

\bibliographystyle{acl_natbib}
\bibliography{emnlp-ijcnlp-2019}

\end{document}